# Characterizing the Temporal, Emotional, and Social Patterns of Adolescent Substance Use Discussions on Reddit

Leran Hong[1], Lei Jin[2], Jianfeng Zhu[3]

[1] Shenzhen Chinese International College, Shenzhen, China

[2] Department of Information Systems and Business Analytics, Kent State University, Kent, OH USA

[3] Departments of Computer[1], Kent State University, Kent, OH USA

## Abstract

Adolescence is a critical developmental period marked by heightened emotional sensitivity, social stress, and vulnerability to substance use. However, traditional research methods provide limited access to adolescents' authentic experiences, hindering efforts to develop evidence-based prevention and intervention strategies. Social media provides a unique opportunity to observe adolescents' naturally occurring discussions about substance use, offering valuable insights into their opinions, emotions, and lived experiences that can inform early prevention and intervention strategies.

In this study, we analyze large-scale Reddit discussions related to substance use among adolescents between 2018 and 2023. Leveraging hour-by-day temporal analysis, sentiment and emotion classification, and transformer-based topic modeling (BERTopic), we examine the interaction between time, emotion, and semantic content in adolescent substance use discourse. Our findings reveal pronounced weekend and late-night peaks in substance-related discussions, a dominance of negative emotions such as sadness and fear, and distinct semantic topics centered on peer relationships, family conflict, emotional distress, and substance-specific experiences. These findings advance our understanding of adolescent substance use in naturalistic online settings and provide empirical evidence to support the development of more timely, targeted and evidence-based prevention and intervention strategies.

## 1. Introduction

Adolescence is a critical developmental period during which substance use can have lasting consequences for mental health and psychosocial development [1]. Early initiation of substance use has been associated with an increased risk of mental health disorders, poorer educational and social functioning, and the later development of substance use disorders, highlighting the importance of early prevention and intervention [2]. Therefore, understanding the behavioral and emotional contexts surrounding adolescent substance use is essential for developing effective prevention and early intervention strategies.

Traditional approaches to studying adolescent substance use primarily rely on self-report surveys, school-based assessments, interviews, or clinical records. Although these methods provide valuable evidence, they are inherently retrospective and episodic, limiting their ability to capture adolescents' rapidly changing emotional experiences and behavioral patterns over time [3]. Adolescents may also underreport substance use because of stigma, privacy concerns, fear of punishment, or social desirability, further restricting access to authentic experience [4]. Consequently, traditional methods provide only limited insight into when adolescents experience heightened vulnerability, how they emotionally respond to substance-related experiences, and the social contexts in which these experiences occur.

Social media offers a complementary source of behavioral evidence by capturing adolescents' spontaneous expressions of emotions, experiences, and social interactions in naturalistic setting [5]. Online communities allow adolescents to discuss sensitive issues, seek peer support, and share personal experience in relatively anonymous environments [6]. Unlike structured questionnaires, these expressions are spontaneous and often emotionally rich. Importantly, social media data include precise timestamps, enabling fine-grained temporal analyses [7] that can reveal daily and weekly rhythms in adolescent discourse [8]. Social media provides a unique opportunity to observe authentic behavioral signals that are often difficult to capture through conventional research methods.

Recent studies have demonstrated the value of social media for understanding adolescent mental health and substance use. Prior work has used Reddit to examine changes in mental health discussion during COVID-19 [9], characterize how teenager frame substance use [10], and investigate the relation of emotion with contextual environment within substance use related discussion [11]. Despite growing interest in social media–based adolescent substance use research, several gaps remain. First, prior studies have paid limited attention to the timing of adolescents' substance-related posts, leaving unclear when these discussions are most likely to occur and what temporal contexts may be associated with heightened vulnerability. Second, emotional expression is frequently analyzed independently of time, leaving unclear when adolescents are most emotionally vulnerable. Third, many studies rely on keyword-based or lexicon-driven methods that struggle to capture the informal, contextual, and emotionally nuanced language typical of adolescent online communication [12].

This study investigated substance use related posts on Reddit from 2018 to 2023 by applying BERTopic for topic modelling and figuring temporal analysis. Reddit is a large online platform composed of topic-specific communities called subreddits [13]. The study conducts temporal analysis to trace changes in the volume and patterns of discussion over time [14], also applying sentiment and emotion analysis to identify the affective dimensions of these posts [10].To address these gaps, this study analyzes large-scale Reddit discussions related to adolescent substance use collected between 2018 and 2023. Specifically, we investigate three research questions:

RQ1: When are adolescents most likely to discuss substance use online?

RQ2: What emotional characteristics are reflected in these discussions?

RQ3: What behavioral contexts and semantic themes characterize adolescent substance use discourse?

To answer these questions, we integrate temporal analysis, transformer-based sentiment and emotion classification, and BERTopic topic modeling. By examining the temporal, emotional, and semantic characteristics of adolescent substance use discussions, this study provides a more comprehensive understanding of adolescents' online experiences and offers behavioral evidence to support more timely and evidence-based prevention and intervention strategies.

## 2. Related Work

### 2.1 Adolescent Substance Use and Mental Health Risks

Adolescent substance use represents a critical public health issue because of its profound and enduring impact on mental health. Early initiation of substance use is not merely a transient risk behavior; it is strongly linked to the later development of substance use disorders and a range of mental health disorders, including mood and anxiety disorders [15]. In the shorter term, adolescent substance use disorders frequently co-occur with depression, anxiety, suicidality, academic difficulties, and social impairment, creating a complex clinical picture that compromises multiple domains of functioning [16]. Given that adolescence constitutes a sensitive developmental period characterized by ongoing neurobiological and psychosocial maturation, early exposure to substances can derail typical developmental trajectories, highlighting why early prevention and intervention are essential. Therefore, effective prevention requires not only identifying substance use but also understanding the emotional and social contexts in which these behaviors emerge.

### 2.2 Limitations of Traditional Substance Use Surveillance

Traditional surveillance of adolescent substance use typically depends on self-reported surveys, interviews, and clinical records. A fundamental limitation of these methods is that substance use is a sensitive behavior and therefore prone to underreporting [17]. Self-reported substance use can be systematically influenced by underreporting bias, which compromises the accuracy of prevalence estimates [18]. Adolescents, in particular, may avoid disclosing substance use due to stigma, concerns about privacy, fear of punishment, and social desirability bias. Social desirability bias has been shown to affect self-reports of mental health, substance use, and related social network factors [19]. Beyond individual reporting biases, the sampling frames commonly used in school-based surveys introduce further error. Such surveys can miss out-of-school adolescents and those at higher risk, leading to an underestimation of true substance use. Indeed, substance use prevalence surveys contain multiple sources of error, and school-based adolescent surveys may significantly underestimate substance use. Collectively, these limitations highlight the need for complementary behavioral data sources that capture adolescents' spontaneous expressions and real-world experiences beyond traditional surveillance approaches.

**2.3 Social Media as a Data Source for Substance Use and Mental Health**

Social media platforms provide large-scale, naturally occurring, user-generated text that can serve as a valuable complement to traditional surveillance methods [20,21]. Adolescents, in particular, may discuss sensitive topics such as substance use more openly in online communities than they would in formal surveys or clinical interviews, thereby circumventing some of the disclosure barriers associated with stigma and social desirability [10]. Prior studies have used Reddit and other platforms for substance use, cessation, and mental health surveillance. For example, prior research analyzed Reddit substance cessation forums and identified distinct emotional expression patterns associated with abstinence and recovery [20]. Similarly, teenagers' substance-use discussions on Reddit have also been examined in prior research, with findings how emotional frames and the surrounding social and psychological contexts shape these conversations [10].These studies demonstrate that social media provides valuable behavioral evidence for understanding adolescent substance use. However, most existing work has focused on identifying substance-related content or describing emotional characteristics, leaving other important dimensions of adolescent online behavior relatively underexplored.

**2.4 Temporal, Emotional, and Semantic Patterns in Online Substance Use Discourse**

Temporal information embedded in social media has been increasingly used to understand health behaviors and behavioral rhythms. Previous studies have shown that online discussions often exhibit meaningful daily and weekly patterns that reflect offline activities and risk behaviors. For example, crowdsensing and social media studies have identified Friday through Sunday evenings as periods associated with increased alcohol use and problem drinking [22]. These temporal patterns have been linked to reduced adult supervision, increased peer interaction, and greater opportunities for unstructured social activities, all of which are recognized risk factors for adolescent substance use [23]. Emotion analysis has become an important approach for understanding substance-related discourse on social media. Transformer-based language models have enabled more accurate identification of emotional expressions beyond traditional lexicon-based approaches. Previous studies consistently report elevated levels of sadness, fear, anger, guilt, and shame in substance-related discussions, suggesting that online conversations frequently reflect emotional distress and psychological vulnerability [11]. Topic modeling has been widely applied to uncover the major themes underlying online substance use discussions. Recent studies have identified recurring topics involving peer influence, family relationships, school stress, party environments, prescription drugs, and other substance-specific experiences [11,24].

Despite these advances, important research gaps remain. Existing studies have generally examined temporal patterns, emotional characteristics, and semantic topics independently, providing only a partial understanding of adolescent substance use discussions. Moreover, relatively little attention has been paid to when adolescents discuss substance use online and how temporal information complements emotional and semantic analyses. Finally, few studies have integrated fine-grained temporal analysis, transformer-based sentiment and emotion classification, and BERTopic topic modeling within a unified analytical framework for understanding adolescent substance use discourse.

## 3. Data and Methods

### 3.1 Data Collection and Preprocessing

This study analyzed a corpus of 31,724 substance-related posts from Reddit posts collected from r/teenager subreddit [25]. The final dataset was stored in a structured CSV file containing author identifiers, UTC timestamps, post titles, self-texts, and a combined posts field created by concatenating the title and body text for subsequent analyses.

Data preprocessing was performed in Python. Text preprocessing included lowercasing, removal of URLs and punctuation, and filtering of low-information conversational tokens to reduce noise while preserving the semantic content of the posts. The cleaned texts were then retained for subsequent sentiment, emotion, and topic analyses.

### 3.2 Temporal, Sentiment, and Emotion Analysis

Temporal patterns were conducted to examine the daily and weekly patterns of adolescent substance use discussions on Reddit. Because each Reddit post is associated with a precise UTC timestamp, temporal information can be used to identify periods of increased online activity and potential behavioral vulnerability [26]. Post timestamps were converted into hour-of-day and day-of-week variables to examine recurring daily and weekly posting patterns.

Following preprocessing, each post was represented using term frequency-inverse document frequency (TF-IDF), which assigns weights to terms according to their importance within the corpus. These feature representations served as the input for subsequent text classification.

Sentiment classification was implemented as a three-class classification task, categorizing posts as positive, negative, or neutral. Emotion analysis was performed as a separate multiclass classification task, assigning each post to one of seven emotion categories: sadness, fear, anger, joy, shame, guilt, and disgust. Both tasks were implemented using a Linear Support Vector Classifier (LinearSVC), a widely used algorithm for sparse, high-dimensional text classification that has demonstrated strong performance on short social media texts. The resulting sentiment and emotion labels were subsequently used to examine the affective characteristics of adolescent substance use discussions [27].

### 3.3 Topic Modeling

To identify the major semantic themes underlying adolescent substance use discussions, we applied BERTopic, a transformer-based topic modeling framework designed for short-text corpora [28]. BERTopic first converts each Reddit post into dense semantic embeddings using the pretrained all-MiniLM-L6-v2 Sentence

Transformer model [29,30], allowing semantically similar posts to be represented in a shared embedding space These embeddings capture contextual semantic similarity beyond simple keyword co-occurrence [30,31,32].

Compared with traditional topic models such as Latent Dirichlet Allocation (LDA), BERTopic is particularly well suited for Reddit posts because it leverages contextual transformer embeddings to better capture semantic meaning in short, informal, and highly variable user-generated text. After topic generation, the resulting topics were manually reviewed and assigned descriptive labels based on the highest-ranked keywords and representative posts to facilitate interpretation of adolescent substance use discourse.

## 4. Results

### 4.1 Temporal Patterns of Adolescent Substance Use Discussions

Figure 1 illustrates the temporal distribution of adolescent substance use discussions across hours of the day and days of the week. Overall, posting activity exhibited clear temporal regularities rather than occurring uniformly throughout the week. Substance-related discussions were consistently more frequent during weekends, with posting activity increasing from Friday evening and remaining elevated through early Sunday morning.

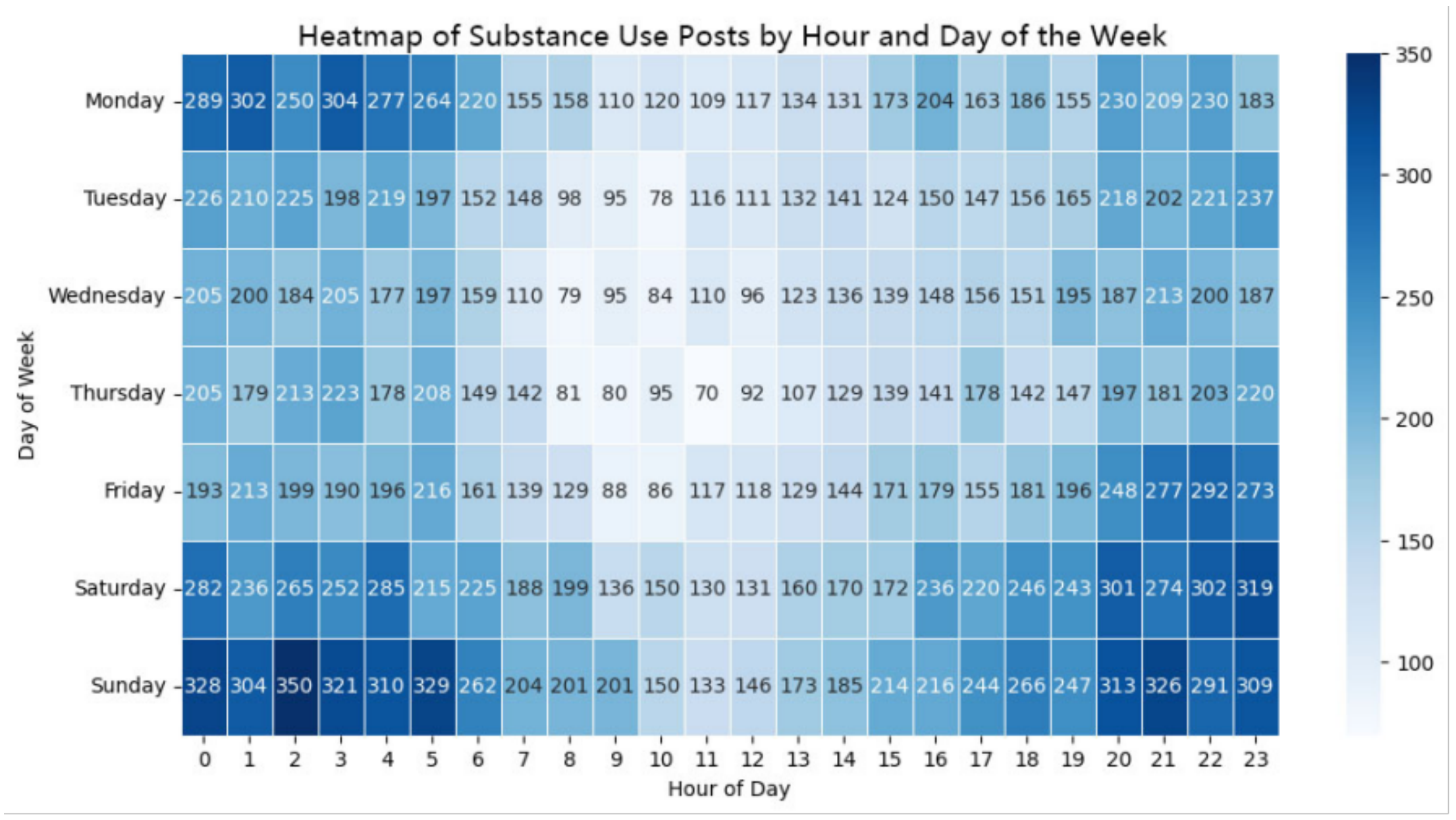

**Figure 1.** Heatmap illustrating the temporal distribution of adolescent substance use discussions on Reddit by hour of the day and day of the week.

A second prominent pattern was the concentration of posts during late-night hours. Across multiple weekdays, posting activity was consistently higher between approximately 9:00 PM and 2:00 AM than during daytime hours, indicating recurring nocturnal peaks in adolescent substance use discussions.

The highest posting intensity occurred on Sunday at approximately 2:00 PM, representing the single peak in discussion volume during the study period. Taken together, these findings demonstrate that adolescent substance use discussions exhibit stable temporal rhythms characterized by increased activity during weekends and nighttime hours rather than occurring uniformly over time.

### 4.2 Sentiment and Emotional Characteristics

Figure 2 (A) demonstrates the distribution of sentiment and emotion categories in adolescent substance use discussions on Reddit. Positive and negative posts accounted for nearly equal proportions of the corpus, representing 47.5% and negative 46.5% of all posts, respectively, whereas neutral posts comprised only 6.0%. The sentiment distribution indicates that adolescent substance use discussions were predominantly associated with positive or negative emotion valence, with relatively few neutral expressions.

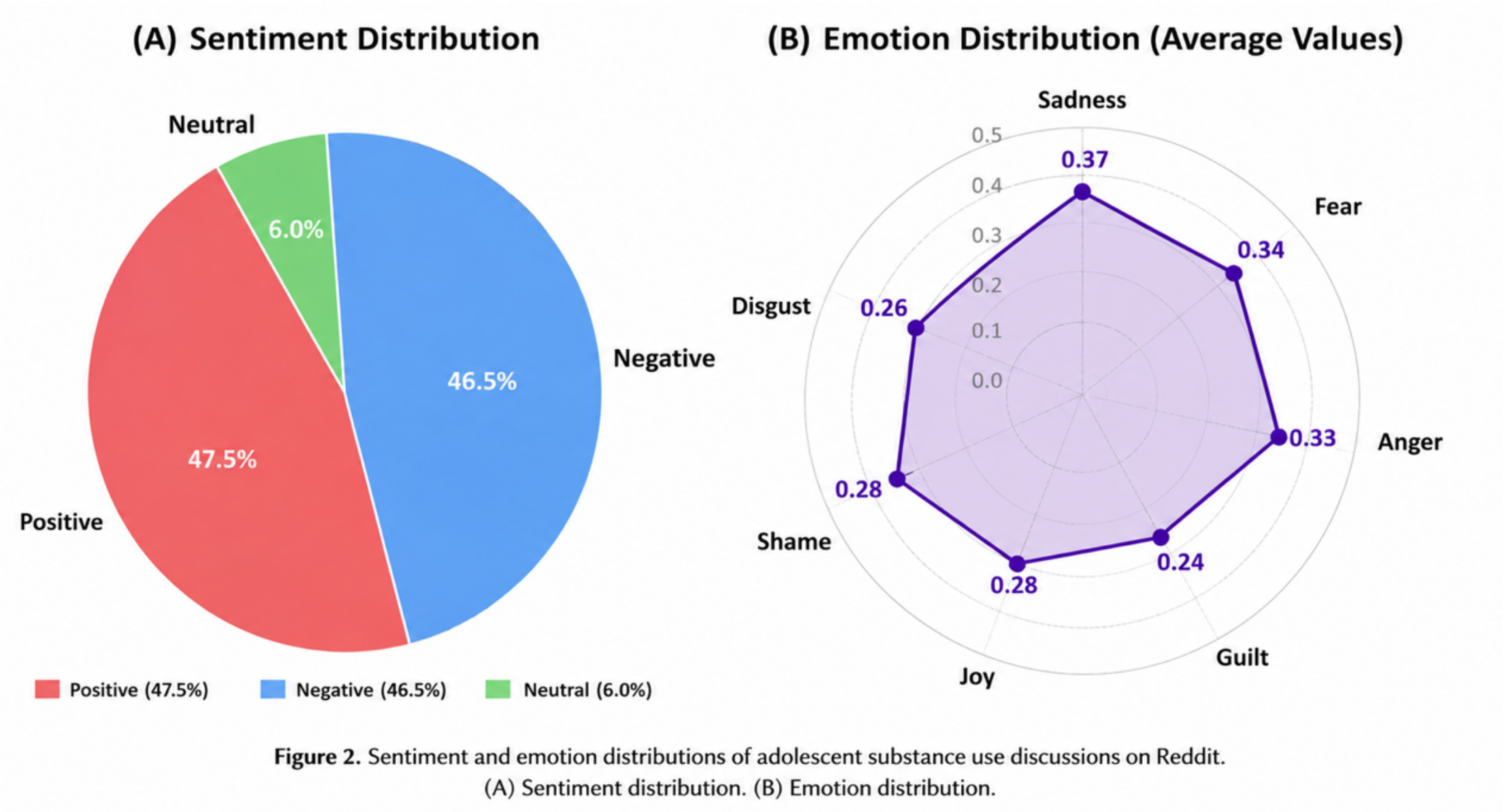

**Figure 2.** Sentiment and emotion distributions of adolescent substance use discussions on Reddit.

Figure 2(B) indicates that the emotional distribution of adolescent substance-use discussions was dominated by sadness and fear, with anger ranking next, while joy, shame, guilt, and disgust showed lower average values. The radar chart demonstrates clear variation across emotion categories, with the largest peaks observed for sadness and fear and smaller values observed for the remaining emotions. Taken together, these results show a concentrated emotional pattern in which a limited number of emotions accounted for most of the affective profile of the corpus.

Figure 3 illustrates the distribution of post lengths across positive, negative, and neutral sentiment categories. Posts expressing positive or negative sentiment tended to be longer than neutral posts. Positive and negative posts exhibited broader length distributions, whereas neutral posts were concentrated at shorter lengths. The average post length across the corpus was 319.19 words.

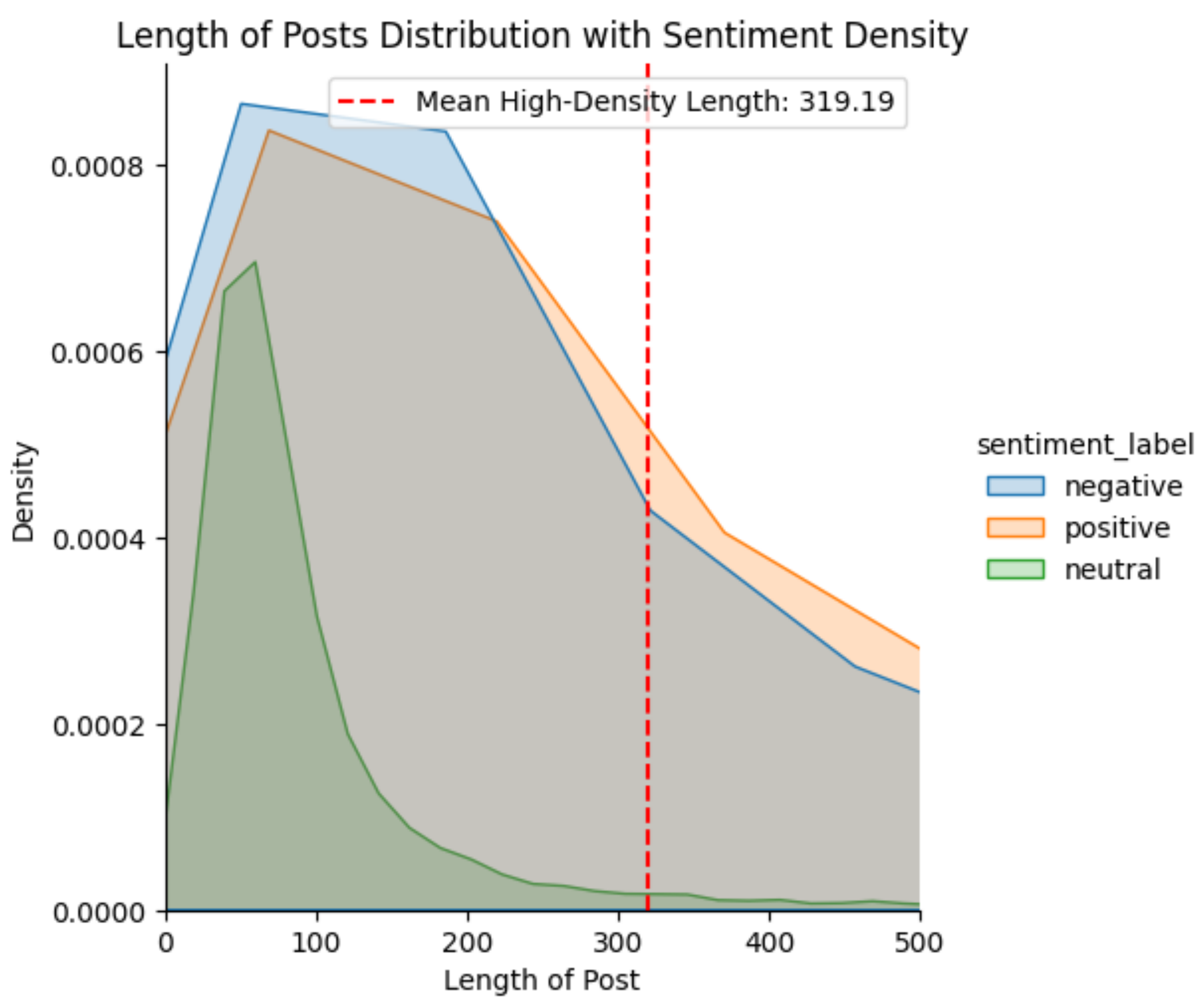

**Figure 3.** Distribution of post lengths across sentiment categories.

Figure 4 presents the most frequently occurring words in the corpus. Substance-related terms, including weed, drunk, vape, alcohol, heroin, LSD, and cocaine, were the most prominent. In addition, words such as school, social, and driving frequently appeared, indicating that discussions extended beyond substance use to broader social and everyday contexts.

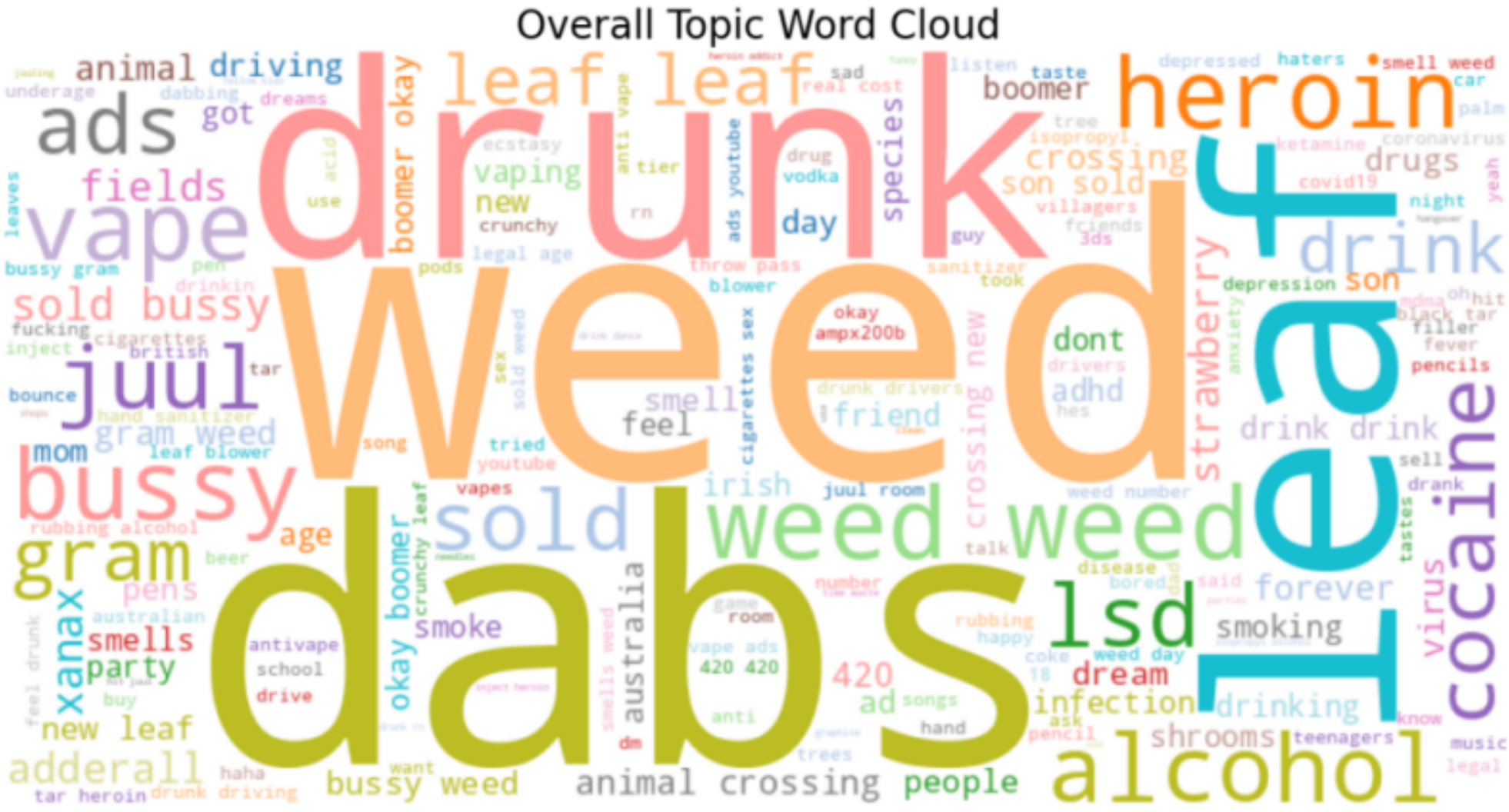

**Figure 4.** Word cloud of the most frequent terms in adolescent substance use discussions on Reddit.

### 4.3 Semantic Topics and Topic Relationships

Table 1 summarizes the eight most prevalent semantic topics identified by BERTopic. The largest topic was Teen Relationship (11.23%), followed by Marijuana Use (10.28%), Alcohol Issues (6.59%), and Teen Vaping (5.99%). Other recurring topics included Party Drinking, Gang Culture, School Life, Cocaine Use, and School Incidents. Overall, the identified topics indicate that adolescent substance use discussions encompassed both substance-specific experiences and broader interpersonal and school-related contexts.

Table 1. Major semantic topics identified by BERTopic and their representative keywords.

| Topic Name | representation | Percentage |
|---|---|---|
| Teen Relationship | Dad, mom, feel, school | 11.23% |
| Marijuana Use | Girl, friend | 10.28% |
| Alcohol Issues | Weed, smoked wee, high, | 6.59% |
| | marijuana | |
| Teen Vaping | Vape, cigarettes, nicotine, smoking | 5.99% |
| Party Drinking | Alcohol, drink, age, buy | 2.45% |
| Gang Culture | Drunk, talk, bored | 2.29% |
| School Life | School, teacher, kid, class | 2.10% |
| Cocaine Use | Weed, mom, parents, dad, smoke | 1.48% |

## 5. Discussion

### 5.1 Overall finding

This study provides a comprehensive characterization of adolescent substance use discussions on Reddit by integrating temporal, affective, lexical, and semantic analyses. Three major findings emerged. First, adolescent substance use discussions exhibited clear temporal regularities, with posting activity peaking during weekends and late-night hours, suggesting that these discussions occurred during specific periods rather than being evenly distributed over time. Second, sentiment and emotion analyses revealed that substance-related discussions were predominantly emotionally charged, with positive and negative sentiment accounting for nearly all posts and sadness, fear, and anger representing the most prominent emotional states. Finally, semantic topic modeling identified multiple coherent discussion themes centered on interpersonal relationships, marijuana and alcohol use, vaping, school experiences, and family contexts, indicating that adolescent substance use discussions extend beyond substance-specific behaviors to broader social and developmental experiences. Collectively, these findings demonstrate the value of social media analytics for providing a multidimensional understanding of adolescent substance use and offer behavioral evidence that may inform future prevention and intervention efforts.

### 5.2 Temporal Patterns of Adolescent substance use discussion

One of the most notable findings was the clear temporal rhythm of adolescent substance use discussions. Posting activity consistently increased during weekends and late-night hours, periods that coincide with reduced adult supervision and greater opportunities for unstructured peer interaction. These findings are consistent with previous ecological momentary assessment and behavioral studies demonstrating that adolescent alcohol and cannabis use occurs more frequently during weekends, holidays, and evening social gatherings[23]. Likewise, routine activity theory suggests that unstructured social time provides greater opportunities for both substance use and discussions surrounding these experiences [33].The observed temporal patterns suggest that online discussions may reflect the same behavioral windows during which adolescents are more likely to encounter substance-related situations in everyday life [34]. Consequently, social media may provide a useful source of real-time behavioral information for identifying periods during which prevention messages, outreach, or digital interventions could be most effectively delivered.

**5.3 Emotional Characteristics within Social Contexts**

The emotional patterns identified in this study further support the social nature of adolescent substance use discussions. Sadness and fear represented the dominant emotions across the corpus, suggesting that many discussions occurred within contexts of emotional distress, interpersonal conflict, or uncertainty. These findings are consistent with previous studies reporting elevated negative affect in online substance use discussions.

Interestingly, positive sentiment accounted for nearly half of all posts despite the predominance of negative emotions. This pattern suggests that positive emotional expressions should not necessarily be interpreted as indicators of psychological well-being. Instead, they may reflect supportive peer interactions, shared experiences, recovery discussions, or the normalization of substance use within online communities. Interpreting emotional expressions alongside semantic topics therefore provides a more nuanced understanding of adolescent substance use than examining emotions alone.

### 5.4 Social Contexts of Adolescent Substance Use

Topic modeling demonstrated that adolescent substance use discussions were strongly embedded within interpersonal and social contexts rather than focusing solely on substances themselves. Family-related topics were among the most prevalent, highlighting the importance of parent-child relationships, family conflict, and communication. Previous research consistently identifies poor family functioning, limited parental monitoring, and parent-child conflict as major risk factors for adolescent substance use, whereas supportive family relationships serve as protective factors [44,45]. The prominence of family-related discussions therefore suggests that adolescents frequently interpret substance-related experiences within broader family relationships.

Peer relationships and party-related discussions further emphasize the social nature of adolescent substance use. Prior research has shown that peer norms, friendship networks, and romantic relationships strongly influence adolescent alcohol and drug use through socialization, identity development, and perceived social acceptance. Likewise, party environments provide highly visible social settings where substance use is normalized and reinforced by peer behavior [35,36]. The prominence of these topics in Reddit discussions suggests that adolescents often describe substance use within broader experiences of belonging, friendship, and social interaction rather than as isolated individual behaviors.

Together, these findings indicate that adolescent substance use should be understood within interconnected family, peer, school, and social environments. Social media discussions therefore provide valuable behavioral evidence for understanding the everyday contexts in which substance use is experienced and discussed.

## 6. Limitations and Future Work

This study has several limitations. First, Reddit users are not representative of the broader adolescent population. The platform is predominantly used by individuals from North America and Europe and tends to overrepresent users with higher levels of internet access, literacy, and socioeconomic resources. Consequently, our findings may not generalize to adolescents from other cultural, geographic, or socioeconomic backgrounds.

Second, social media posts represent self-expressions rather than clinically verified substance use behaviors. Users may exaggerate, conceal, or fictionalize their experiences, and online discussions do not necessarily reflect actual substance use or mental health status. Therefore, our findings should be interpreted as patterns of online discourse rather than direct measures of adolescent behavior.

Third, although transformer-based sentiment and emotion classifiers achieve strong performance on social media text, they may misclassify sarcasm, slang, implicit emotions, or rapidly evolving adolescent language. Similarly, topic modeling identifies semantic patterns but cannot establish causal relationships among family dynamics, peer interactions, emotions, and substance use.

Future research should integrate multiple social media platforms, validate online behavioral signals using external epidemiological or clinical data, and examine longitudinal changes in adolescent substance use trajectories. Future studies may also explore multimodal data sources and large language models to better understand the dynamic interactions among social environments, emotions, and substance use behaviors.

## 7. Conclusion

This study provides a comprehensive characterization of adolescent substance use discussions by jointly examining their temporal, emotional, and semantic patterns using large-scale Reddit data. We found that substance-related discussions were concentrated during weekends and nighttime periods characterized by reduced supervision, while topic modeling highlighted family relationships, peer interactions, and party culture as the primary social contexts associated with substance use. Emotion analysis further revealed that these discussions were predominantly characterized by sadness and fear, although positive emotions also emerged in supportive and peer-related contexts.

Rather than focusing solely on detecting substance use, our findings contribute to a broader understanding of the behavioral and social contexts in which adolescents discuss substance-related experiences. These insights may support the development of more timely, context-aware, and developmentally appropriate prevention and intervention strategies, while also demonstrating the value of social media as a complementary source for adolescent behavioral health research.